\documentclass{article}

\usepackage[square,numbers]{natbib}
\usepackage[utf8]{inputenc}

\usepackage[final]{corl_2019} 

\usepackage{float}
\usepackage{caption} 
\usepackage{makecell}
\usepackage{subfiles}
\usepackage{ amssymb }
\usepackage{amsmath,amsfonts,amssymb,amsthm}
\usepackage{mathtools}
\usepackage{commath}
\usepackage[sc,osf]{mathpazo}
\let\norm\undefined 
\DeclarePairedDelimiter\norm{\lVert}{\rVert}
\usepackage{graphicx}
\usepackage{subcaption} 
\title{Deep Unsupervised Common Representation Learning for LiDAR and Camera Data using Double Siamese Networks}

\author{Andreas~Bühler\thanks{The authors contributed equally to this work.}\qquad
        Niclas~Vödisch\footnotemark[1]\qquad
        Mathias~Bürki\qquad 
        Lukas~Schaupp\\
        Autonomous Systems Lab, ETH Zürich
}

\begin{document}
\maketitle


\begin{abstract}
Domain gaps of sensor modalities pose a challenge for the design of autonomous robots. 
Taking a step towards closing this gap, we propose two unsupervised training frameworks for finding a common representation of LiDAR and camera data. 
The first method utilizes a double Siamese training structure to ensure consistency in the results. 
The second method uses a Canny edge image guiding the networks towards a desired representation.
All networks are trained in an unsupervised manner, leaving room for scalability. 
The results are evaluated using common computer vision applications, and the limitations of the proposed approaches are outlined.
\end{abstract}



\section{Introduction}
\label{sec:introduction}

In the last years a lot of research has been done on improving computer vision and LiDAR systems.
Applications range from autonomous cars to unmanned drones.
One of the main areas of development is mapping and localization.
Mainly two sensors are used for this application: LiDARs, which provide highly accurate distance measurements but are still rather expensive, and cameras, which are cheaper but they lack accuracy when used for mapping.
In this work we propose a step towards a new concept of localization, namely finding a common representation of LiDAR and camera data to be used with standard computer vision algorithms. 
This common representation provides a useful application in swarm robotics like swarm drones or fleets of autonomous cars.
The idea is the following: A mapping device is equipped with one highly accurate LiDAR sensor and utilized to create a precise and consistent 3D~model of the environment, the map.
The swarm robots are equipped with an affordable, lightweight camera. 
The images taken by the camera can be transformed into the proposed common representation, together with nearby point cloud data of the map.
Next, a standard 2D-2D feature matching is performed using the common representations.
The 3D positions of the matched features are retrieved from the stored point cloud based map data, which was originally used to generate one of the common representations.
Using this data, the 6DoF pose of the robot can be estimated. 

\begin{figure}[t]
    \centering
  \begin{subfigure}{\textwidth}
  \centering
    \includegraphics[width=0.8\textwidth]{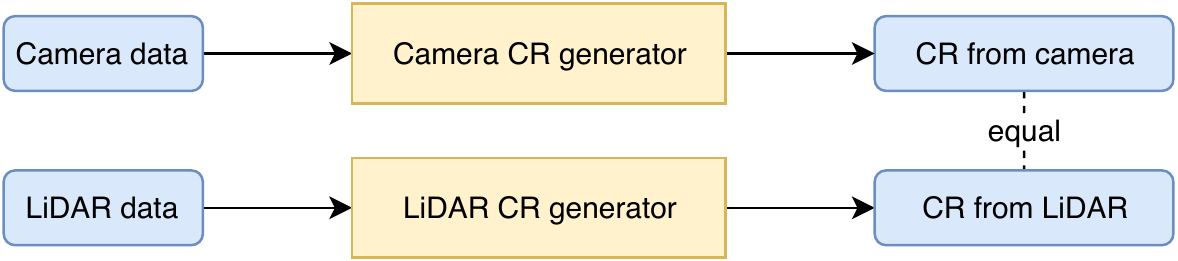}
  \end{subfigure}
  \caption{The common representation (CR) generators leverage the generalization power of neural networks to find a 2D image, the common representation, that can be created from either camera or LiDAR data.}
  \label{fig:high_level_overview}
\end{figure}

Our contribution to this concept are two methods for finding a common representation of image and point cloud data in the form of a 2D image.
The process of finding this common representation is learning based and unsupervised. 
Figure~\ref{fig:high_level_overview} provides a high-level overview of the system.

\begin{figure}[b]
	\centering
  \begin{subfigure}[b]{0.24\textwidth}
    \includegraphics[width=\linewidth]{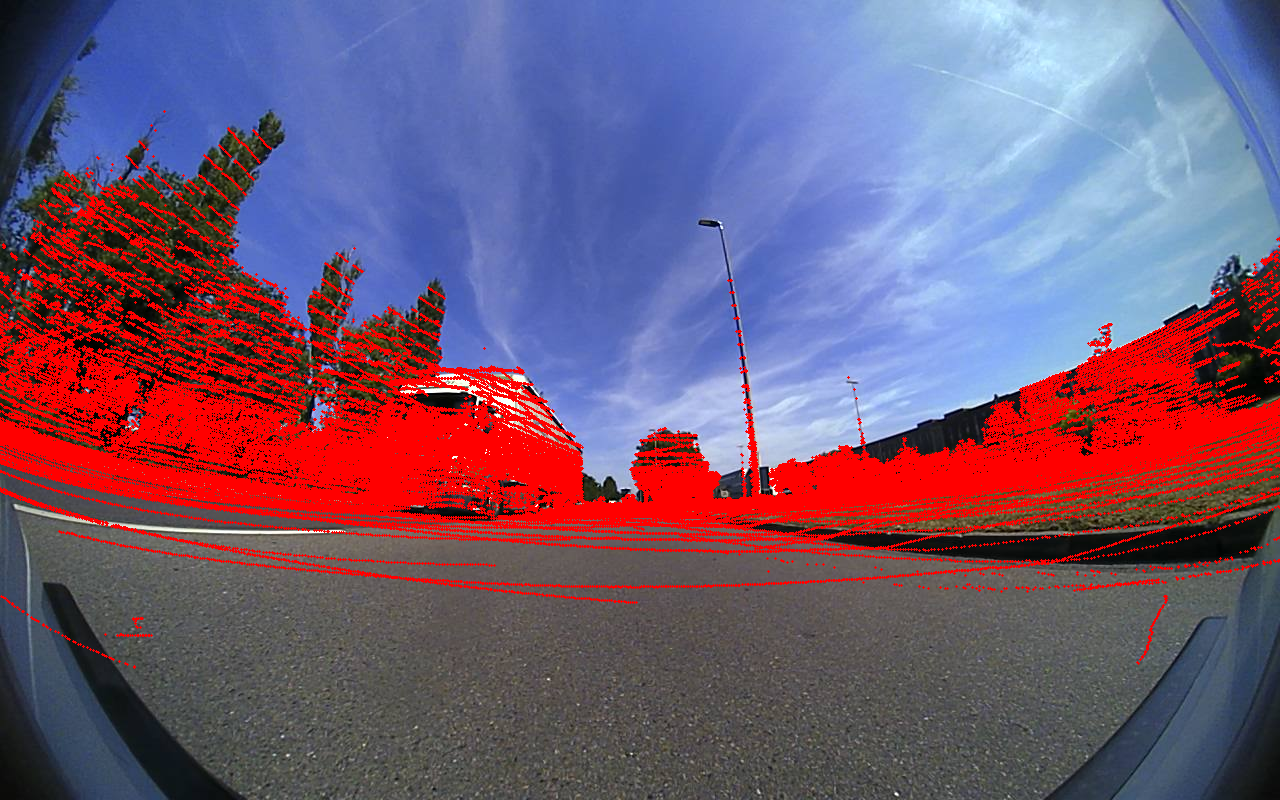}
    \caption{}
    \label{fig:intro-overlay-dist}
  \end{subfigure}
  \begin{subfigure}[b]{0.24\textwidth}
    \includegraphics[width=\textwidth]{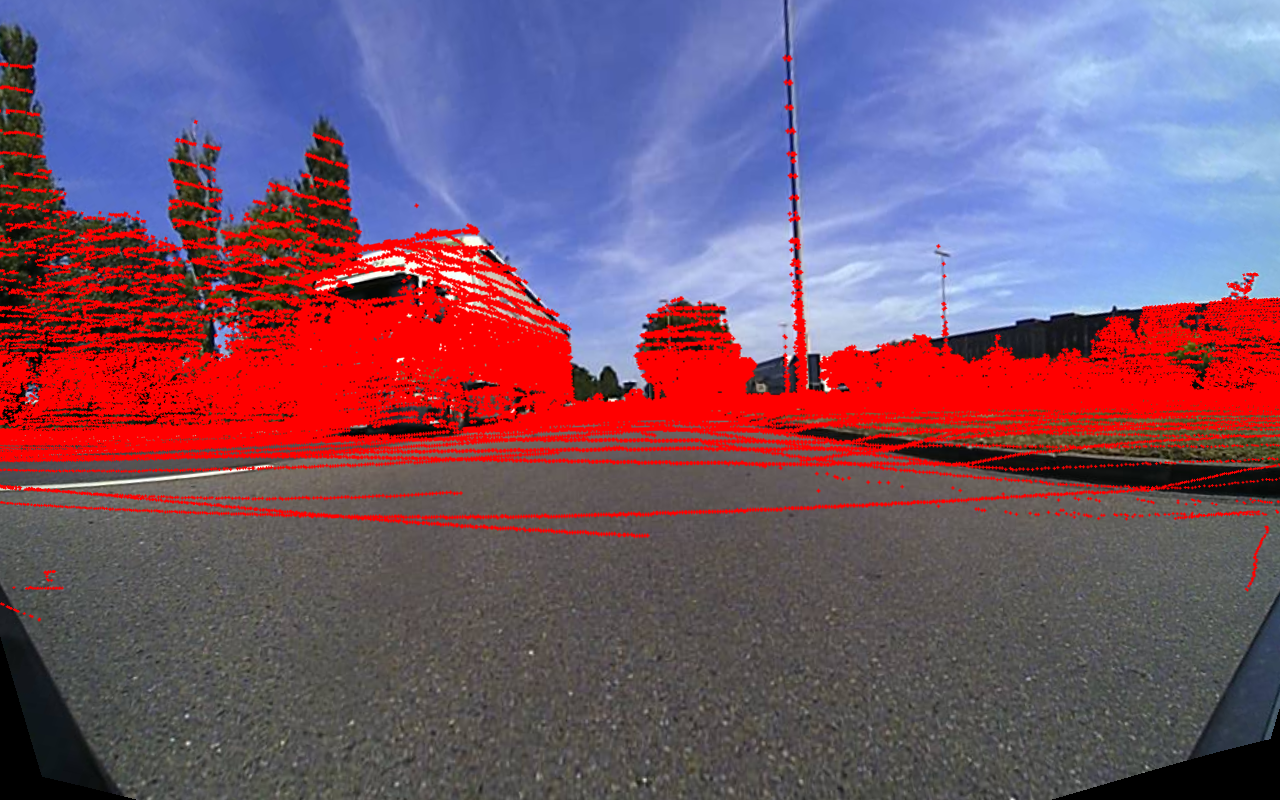}
    \caption{}
    \label{fig:intro-overlay-undist}
  \end{subfigure}
  \begin{subfigure}[b]{0.24\textwidth}
    \includegraphics[width=\textwidth]{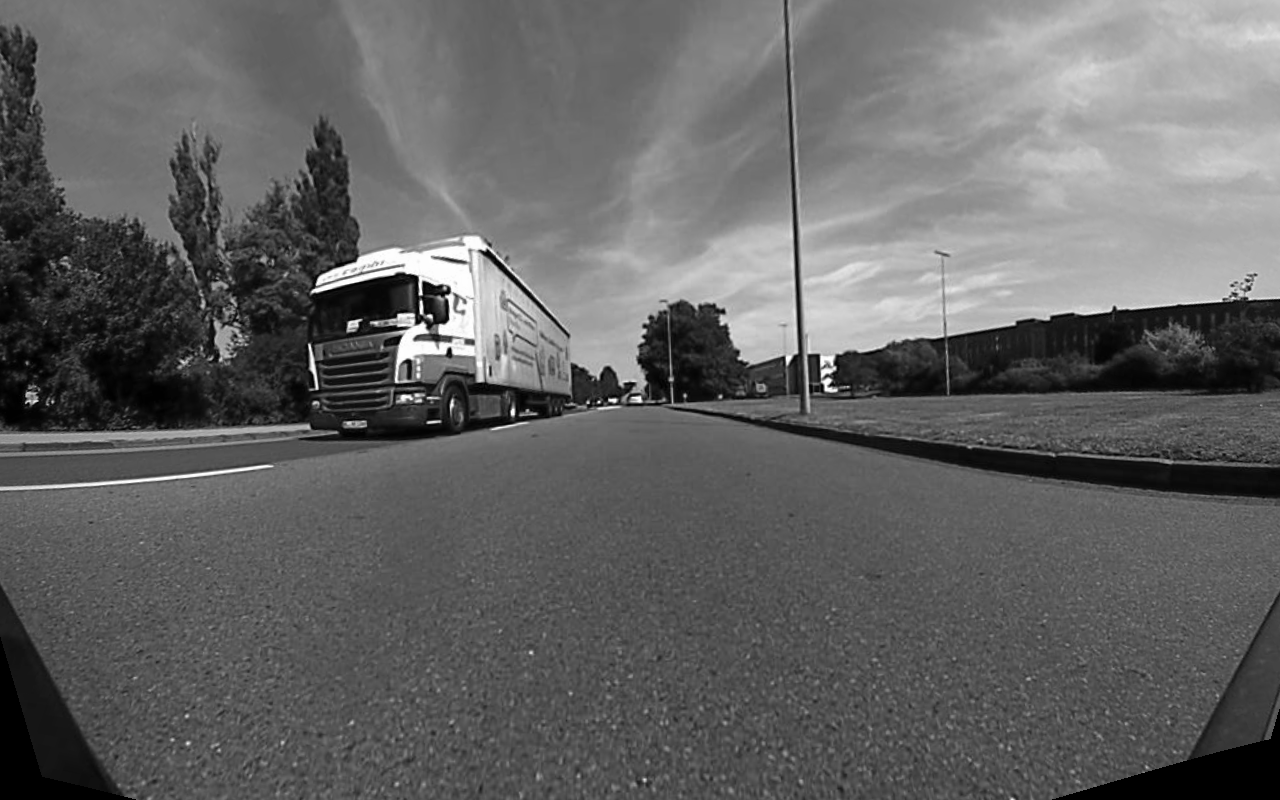}
    \caption{}
    \label{fig:intro-input-img}
  \end{subfigure}
    \begin{subfigure}[b]{0.24\textwidth}
    \includegraphics[width=\textwidth]{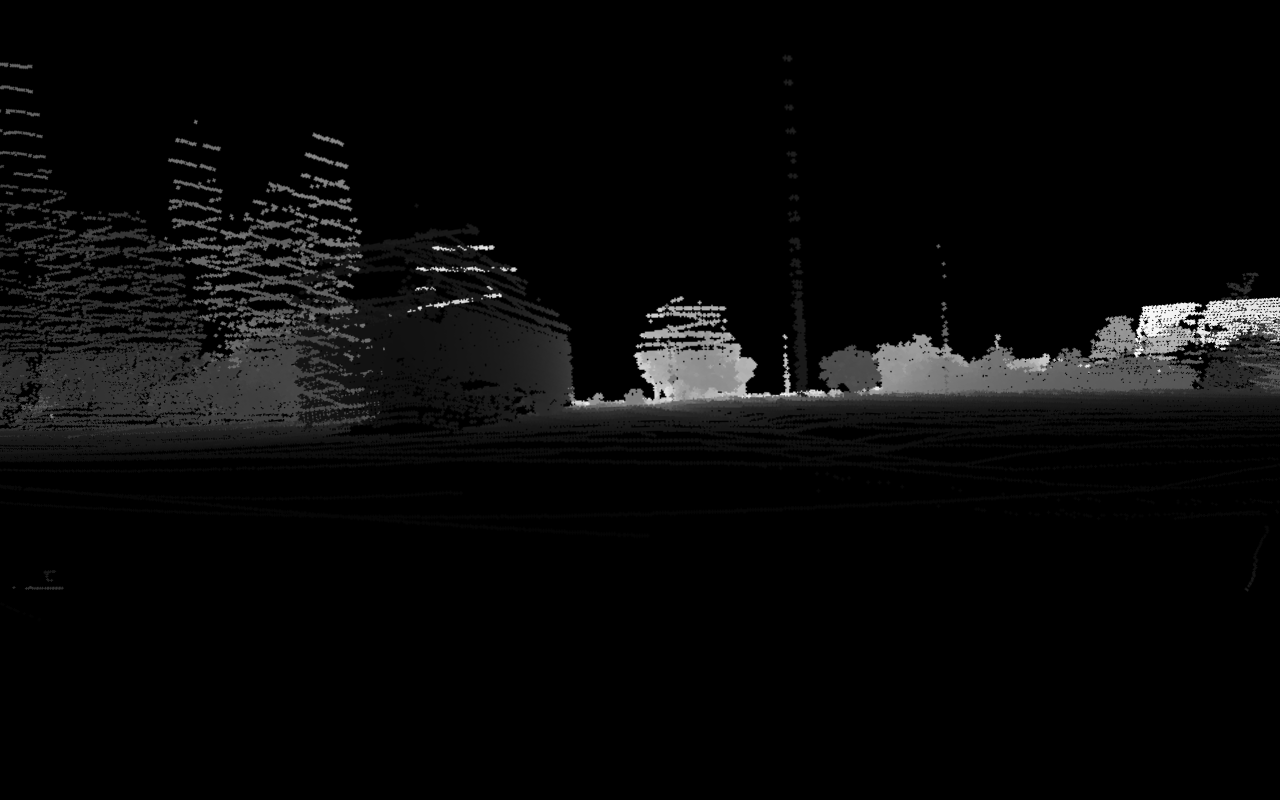}
    \caption{}
    \label{fig:intro-input-pc}
  \end{subfigure}
  \caption{(a) Overlay of point cloud (red) and fisheye distorted image. (b) Overlay of point cloud (red) and image after undistortion. (c) Undistorted, grayscale input image to the proposed networks. (d) Undistorted depth input image to the proposed network, generated by a 2D projection of the point cloud data.}
  \label{fig:intro}
\end{figure}

Figure~\ref{fig:intro} shows the used input data. 
A point cloud created by multiple LiDAR sensors is projected onto the corresponding image of a forward facing camera with a fisheye lens.

\section{Related Work}
\label{sec:related_work}

Finding an abstract representation of images is an extensively studied topic in the field of computer vision and deep learning.
A very common approach to this problem is the utilization of convolutional layers combined in an encoder-decoder network~\citep{Badrinarayanan2017}.
The encoder aims to learn an encoding of the input in a high dimensional hyperspace, while the decoder ensures that this code contains all relevant information to generate the desired output image.
The simplest example are auto-encoders, which reconstruct the input image from the intermediate code.
Those networks are often used as the backbone architecture of more complex structures~\citep{Zeng2017}.
By constraining the learned encoding, variational auto-encoders (VAE) are restricted to include certain information in the low level representation~\citep{Kulkarni2015}.

Another approach to enforce a specific representation of the output leverages the feedback loop in generative adversarial networks (GANs).
A GAN is composed of two subnetworks: while the generator creates artificial data, the discriminator tries to differentiate them from another data set that represents the target domain.
Recent research based on GANs has shown astonishing results when converting images from one space to another space in an unsupervised fashion~\citep{Zhu2017}.
In both~\citep{Choi2017} and~\citep{Kim2017} multiple encoding-decoding networks are coupled to transfer images between several domains where no trivial relations exist.

In order to find a common structure between data taken from multiple modalities, it is important to guide the network during the training process by providing both similar and non-similar pairs of the respective domains.
Siamese networks consist of two networks that share all weights and that are trained simultaneously.
A final cost layer computes the similarity score between the output of the different input streams.
In~\citep{Chung2017} Siamese networks are used to identify persons in images taken by different cameras and across time.
Another field of application is image matching when dealing with changing weather conditions or diverse viewpoints~\citep{Melekhov2016}.

An example of combining the aforementioned work is presented in~\citep{Larsen2015}.
Using output data from a VAE as first input for a GAN's discriminator and having a traditional auto-encoder as the second stream, the encoding learned by the auto-encoder is constrained to contain the desired information.
This overcomes the problem of handcrafted features based on a pixel level.
However, this approach cannot be directly applied to data from multiple modalities.
In SuperPoint~\citep{DeTone2017} the idea of Siamese training is applied to find keypoint correspondences between images in an completely unsupervised fashion.
Classical computer vision methods, e.g., SIFT features, fall short behind the capabilities of SuperPoint.
Even though the visual results are very promising, the method requires the computational power of recent GPU models, which conflicts with equipping low-cost swarm robots.


\section{Method}
\label{sec:method}

In this section two approaches for finding a common representation (CR) between camera and LiDAR data are proposed.
The first approach (section~\ref{subsec:method_siamese})  is purely unsupervised and utilizes a double Siamese network architecture.
The second approach (section~\ref{subsec:method_edges}) tries to find a CR that resembles an edge image, to facilitate the learning process.
Lastly, in section~\ref{subsec:method_layers} the network layers utilized in the proposed architectures are explained.

\begin{figure}
    \centering
  \begin{subfigure}{\textwidth}
  \centering
    \includegraphics[width=0.9\textwidth]{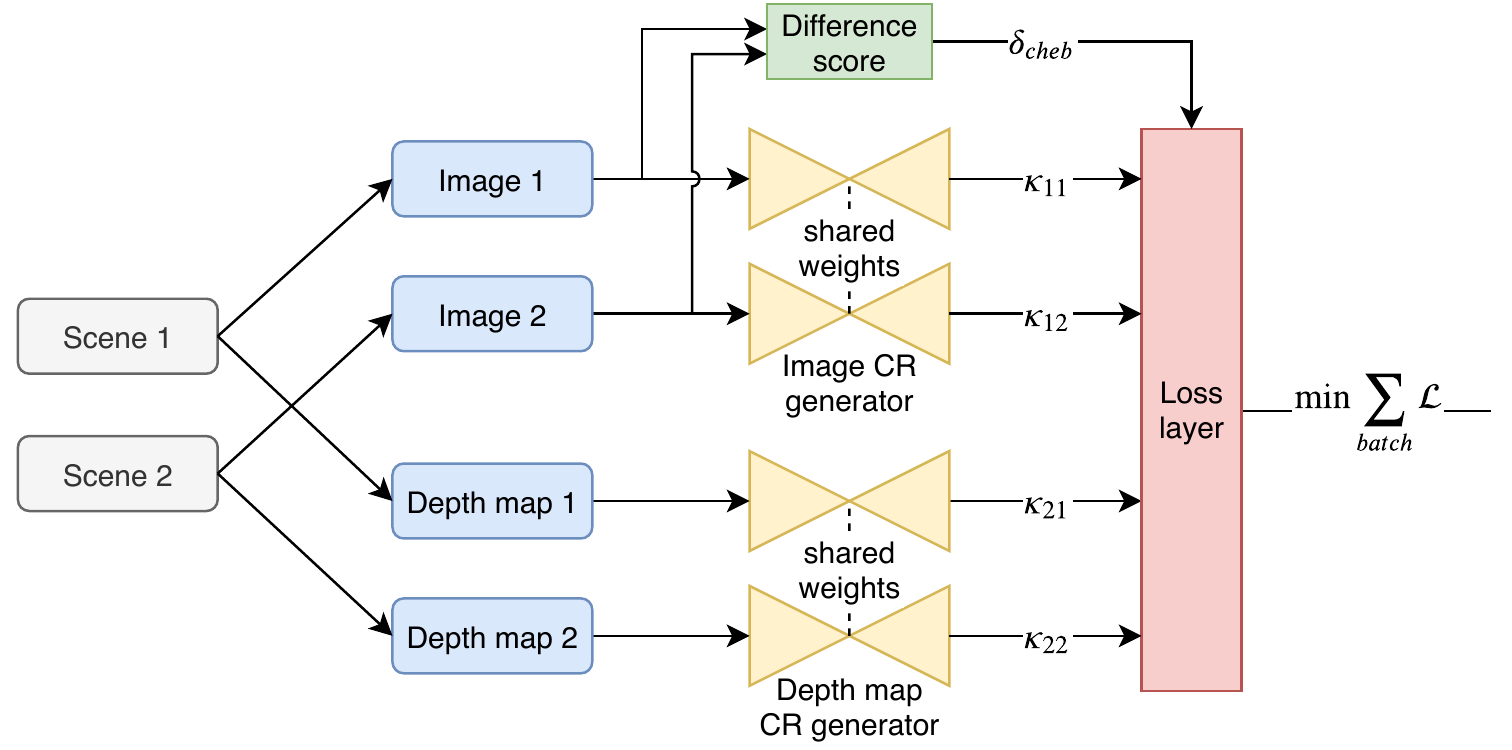}
  \end{subfigure}
  \caption{Double Siamese Networks based on two encoder-decoder subnetworks to generate the CR from two pairs consisting of an image and the corresponding depth map.}
  \label{fig:nn_architecture_colors_common_rep}
\end{figure}

\subsection{Architecture 1 - Double Siamese Networks}
\label{subsec:method_siamese}

Figure~\ref{fig:nn_architecture_colors_common_rep} shows the proposed architecture to learn a CR of image and point cloud data in a purely unsupervised manner.
The framework builds on ideas from Siamese learning and encoder-decoder architectures, which are commonly used in unsupervised approaches.
The Siamese learning approach ensures that the results are consistent, i.e., changes of the input result in similar changes of the output. 
Firstly, two images (image~1 and image~2) are sampled from the database and used to compute a difference score.
This metric describes the similarity between both images and returns a high value for samples of the same scene of slightly different viewpoints.
The Chebyshev distance~\citep{Guo2017} is utilized as it provides the desired behavior. 
During the sampling process, a tuning parameter is used as the probability whether to randomly sample from a set of similar or non-similar images.
Secondly, the system fetches the corresponding depth maps (depth map~1 and depth map~2) belonging to image~1 and image~2, respectively.
The depth maps are computed by projecting the point cloud data onto the same image plane as given by the corresponding images.
The pixels encode the depth measurement of the respective point.
Finally, image~1 and image~2 are fed into the image CR generator, whereas depth map~1 and depth map~2 are fed into the depth map CR generator.
Both generators are designed in an encoding-decoding fashion, which is explained in section~\ref{subsec:method_layers}.
The four forward passes through the subnetworks generate four grayscale output images representing the respective CR of the input pair~1 and input pair~2.
The double Siamese networks are trained using the following loss~$\mathcal{L}$:

\begin{align}
\begin{split}
\mathcal{L}(\kappa_{11}, \kappa_{12}, \kappa_{21}, \kappa_{22}, \delta_{cheb}) = \norm{\kappa_{11}-\kappa_{21}} +
\norm{\kappa_{12}-\kappa_{22}} \\ 
+ \abs{\norm{\kappa_{11}-\kappa_{12}} - \delta_{cheb}} \qquad  \\
+ \abs{\norm{\kappa_{21}-\kappa_{22}} - \delta_{cheb}} \qquad 
\end{split}
\end{align}

where $\kappa_{11}$ and $\kappa_{12}$ are the CRs generated from the first and second image, respectively.
$\kappa_{21}$ and $\kappa_{22}$ are the CRs generated from the first and second depth map, respectively.
$\delta_{cheb}$ is the Chebyshev distance of images 1 and 2, and the norm $\norm{\cdot}$ can theoretically be any matrix/image comparison norm.
To this end the Manhattan distance is utilized.
The intuition behind this formulation of the loss function is the following: The loss should be small when the CR computed from the image and the depth map is similar and, additionally, one of the two following conditions holds: either when the Siamese pairs are similar, i.e., a small difference score, and the CR of the respective Siamese pairs are similar, or when the Siamese pairs are not similar, i.e., large difference score, and the CR of the respective Siamese pairs differ a lot.
Similarly, the loss should be large in the respective inverse cases.


\subsection{Architecture 2 - Common Edges}
\label{subsec:method_edges}

\begin{figure}[H]
  \begin{subfigure}{\textwidth}
  \centering
    \includegraphics[width=0.7\textwidth]{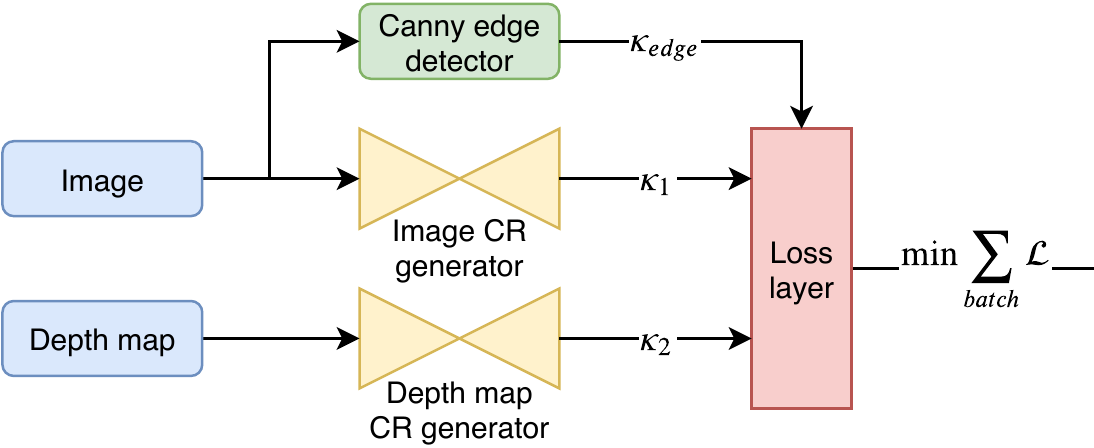}
  \end{subfigure}
  \caption{Common Edges architecture utilized to train a common representation that resembles a Canny edge image.}
  \label{fig:nn_architecture_colors_common_edges}
\end{figure}

Our second approach of finding a common representation utilizes a Canny edge image in the loss function to solve the consistency problem. 
Figure~\ref{fig:nn_architecture_colors_common_edges} shows the proposed architecture of this approach. 
The input, which is randomly fetched from a database, consists of a grayscale image and the corresponding depth map.
The Canny edge algorithm is utilized to generate an edge image based on the grayscale input image. 
The inputs are fed through two separate encoder-decoder networks, which are explained in more detail in section~\ref{subsec:method_layers}.
The following loss function is utilized:

\begin{align}
\begin{split}
\mathcal{L}(\kappa_{1}, \kappa_{2}, \kappa_{edge}) = \norm{\kappa_{1}-\kappa_{2}}
+ \norm{\kappa_{1}-\kappa_{edge}}  
+ \norm{\kappa_{2}-\kappa_{edge}} 
\end{split}
\end{align}

where $\kappa_{1}$ and $\kappa_{2}$ are the CRs generated from the image and the depth map, respectively, and $\kappa_{edge}$ resembles the edge image. 
The norm $\norm{\cdot}$ operator is implemented as the Manhattan distance, but other matrix/image comparison norms could work as well.
This approach ensures that the generated CRs are consistent and resemble an edge image. 
Therefore, a change in the input image leads to a similar change in the output image.

\subsection{Encoder-Decoder Layers}
\label{subsec:method_layers}

\begin{figure}[t]
\centering
  \begin{subfigure}{\textwidth}
    \includegraphics[width=1.0\textwidth]{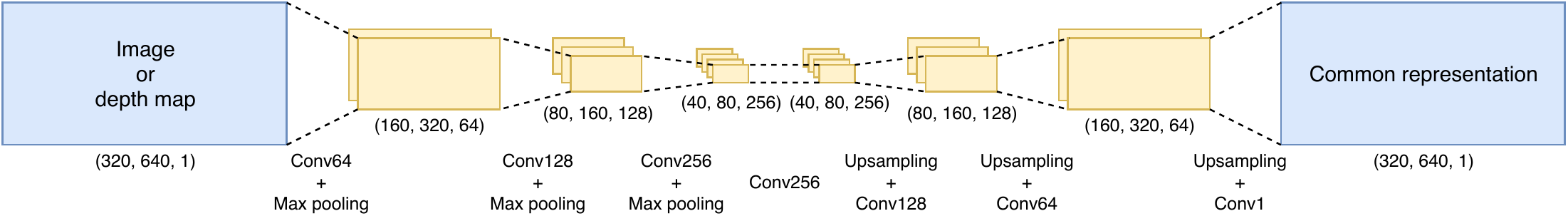}
  \end{subfigure}
  \caption{Network layers used as CR generator in both approaches.}
  \label{fig:nn_architecture_colors_layers}
\end{figure}

Figure~\ref{fig:nn_architecture_colors_layers} depicts the encoder-decoder network architecture that is utilized in the Double Siamese Networks, section~\ref{subsec:method_siamese}, and in Common Edges, section~\ref{subsec:method_edges}.
The input into this network could be any image that fits the input dimension.
For our purposes, the input is either an image or a depth map, and the output is a 2D image, the CR. 
The layers are chosen such that the network learns a higher dimensional representation of the input data in some latent space by extracting the relevant features in an unsupervised fashion.
Since the CR should preserve local information with respect to the input data, the network consists only of convolutional layers.
After each convoulutional layer, a max pooling operation reduces the first two dimensions by a factor of two.


\section{Results}
\label{sec:results}

\begin{figure}[b]
	\centering
  \begin{subfigure}[b]{0.49\textwidth}
    \includegraphics[width=\linewidth]{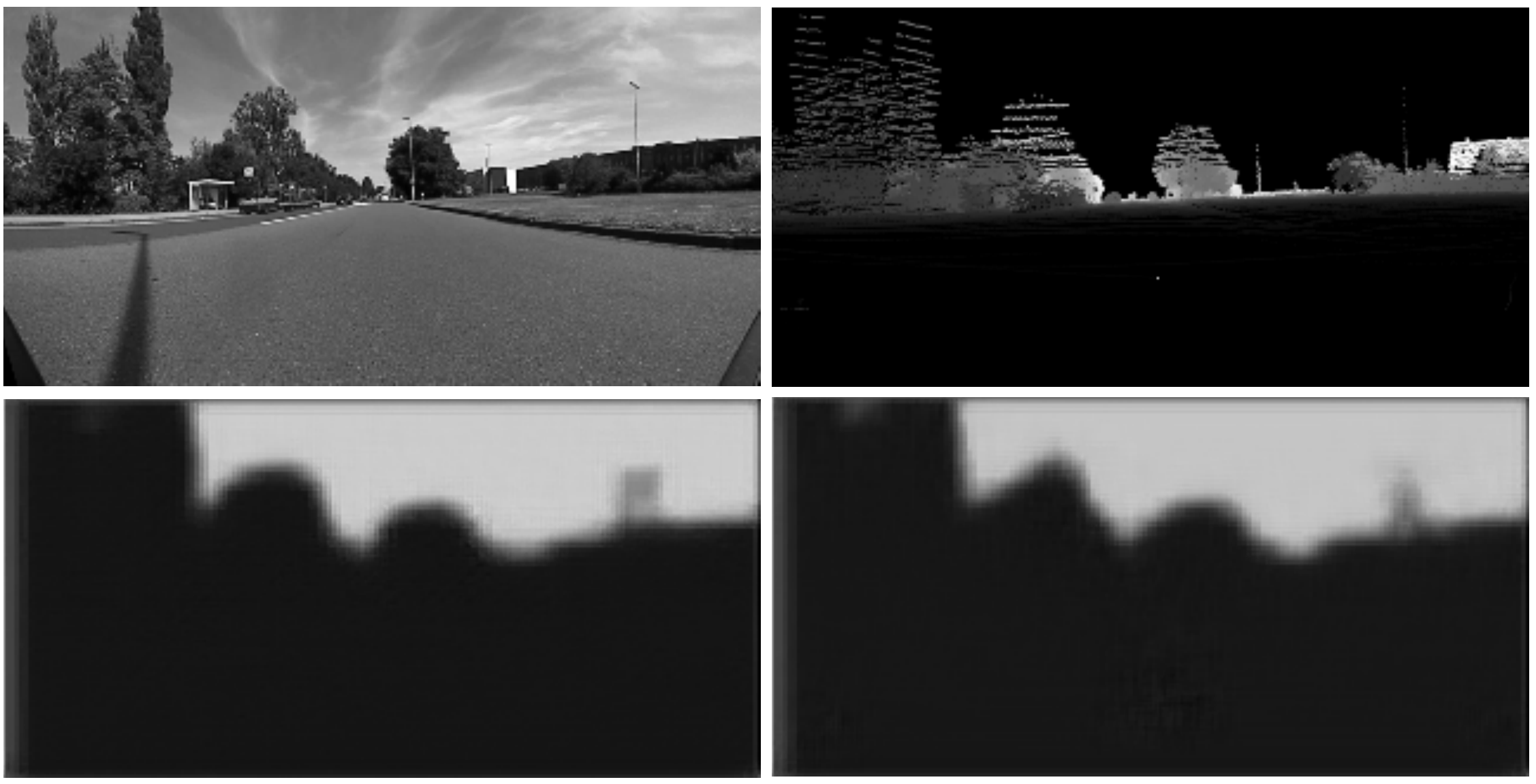}
  \end{subfigure}
  \begin{subfigure}[b]{0.49\textwidth}
    \includegraphics[width=\textwidth]{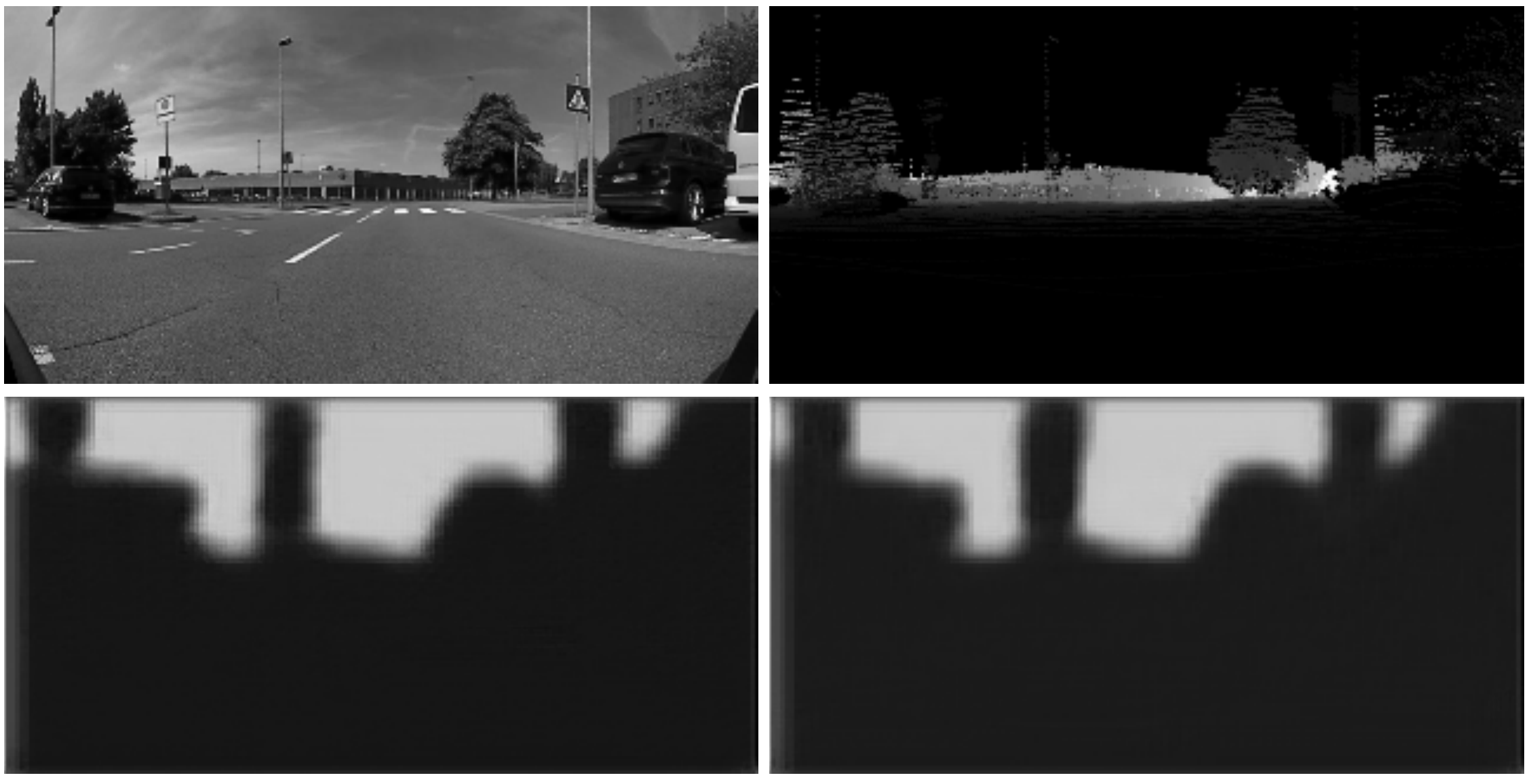}
  \end{subfigure}
  \begin{subfigure}[b]{0.49\textwidth}
    \includegraphics[width=\textwidth]{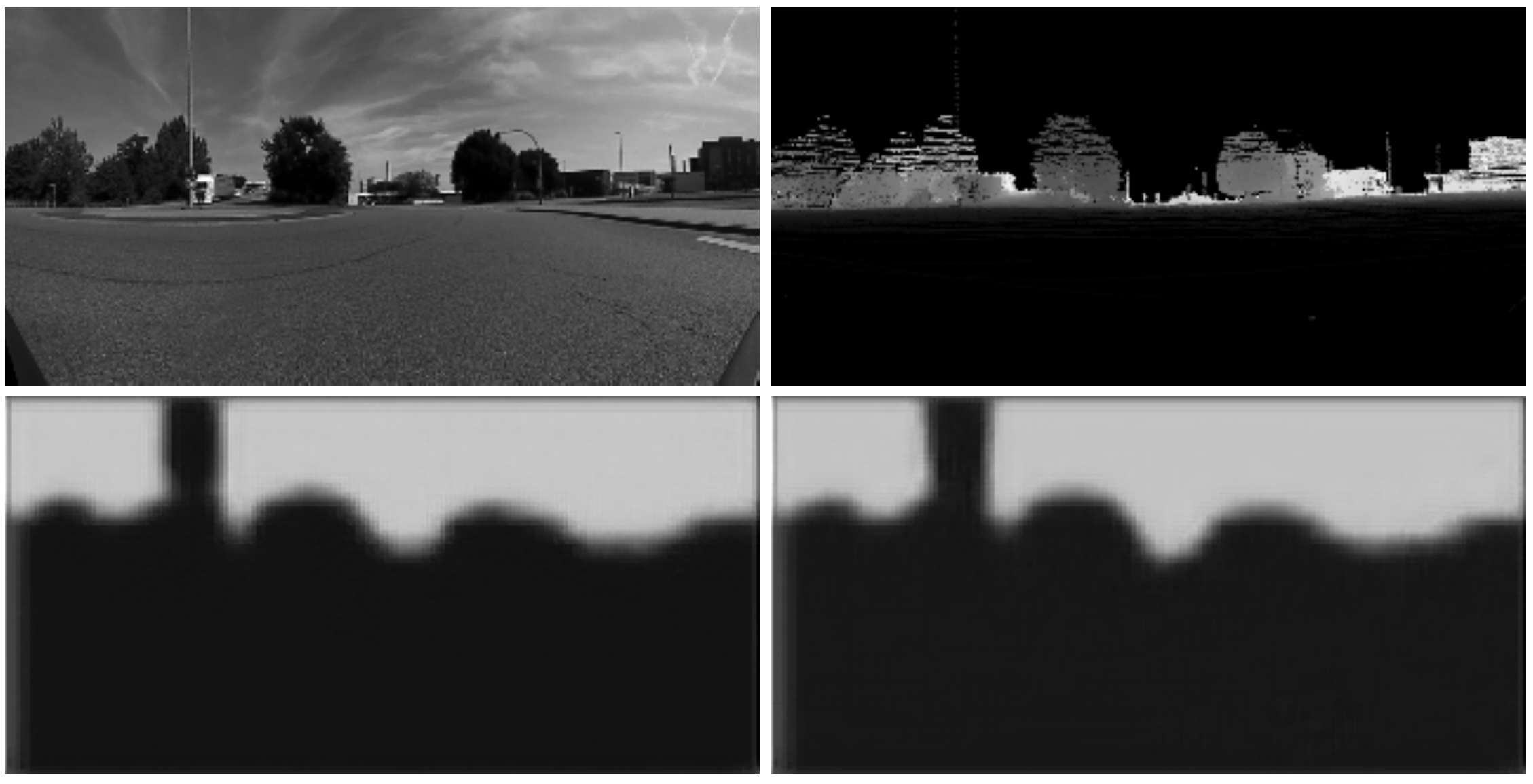}
  \end{subfigure}
    \begin{subfigure}[b]{0.49\textwidth}
    \includegraphics[width=\textwidth]{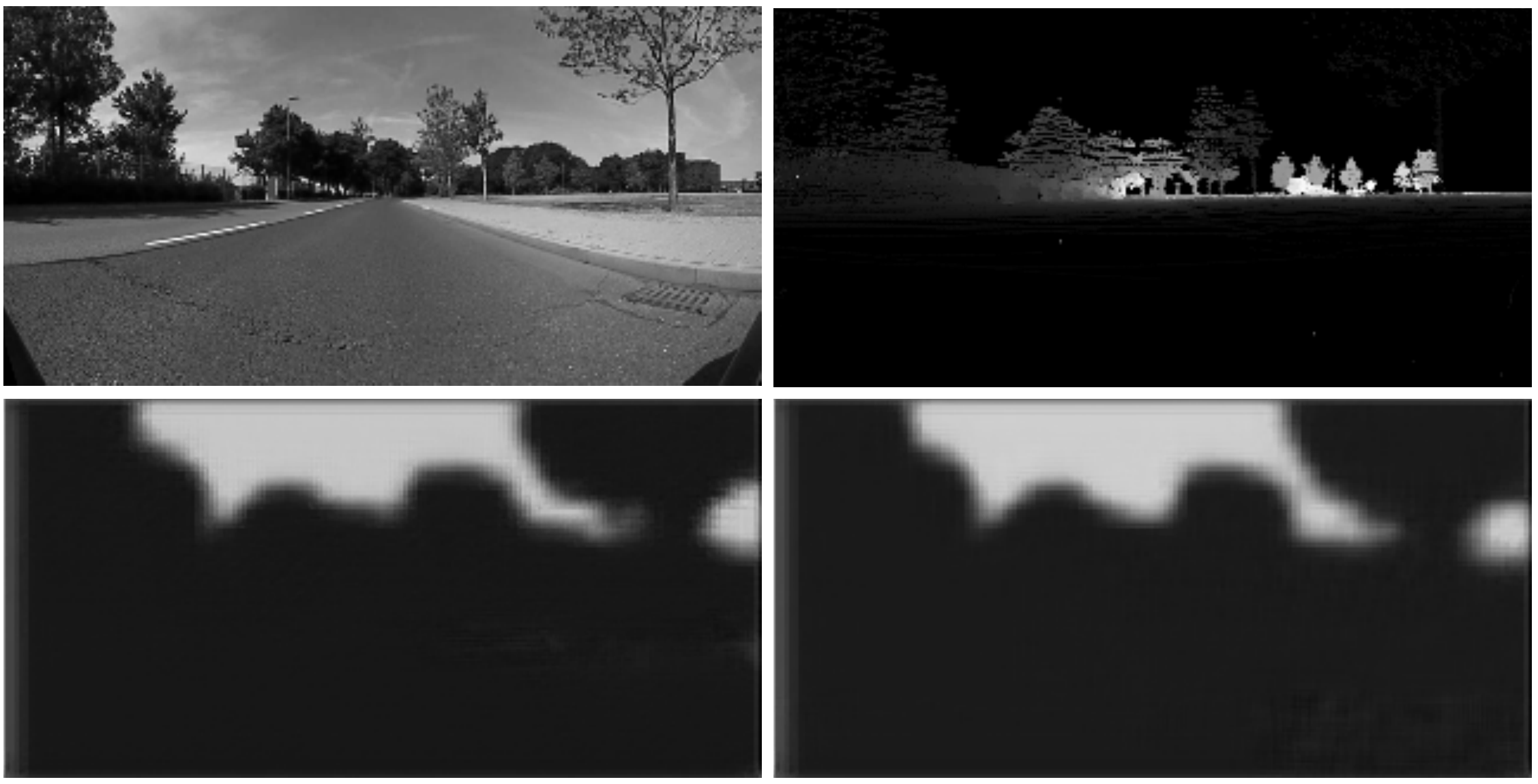}
  \end{subfigure}
  \caption{Top left to bottom right of each group of four images: Input image, input depth image from point cloud, common representation generated from input image, common representation generated from depth image.}
  \label{fig:results_common_rep}
\end{figure}

The proposed architectures are trained on the UP-Drive data set~\citep{Varga2017}. 
The point cloud data is projected onto a 2D plane and utilized as depth image.
The RGB image coming from the front facing camera is converted to grayscale and scaled to a feasible size. 
The dimensions are 320x160 pixels for the Double Siamese Networks and 640x320 pixels for the Common Edges method, facilitating the training process. 

\subsection{Architecture 1 - Double Siamese Networks}
\label{subsec:results_siamese}

Figure~\ref{fig:results_common_rep} shows the results generated by the Double Siamese Networks. 
It can be seen that the CR is very similar when generated from either the grayscale image or depth image, which is what was one of the goals of this work. 
On the other hand, it is is not very descriptive and detailed anymore. 
A lot of information is lost in the compression process of the encoder-decoder networks and due to the large domain gap between the two modalities.

\begin{figure}
	\centering
  \begin{subfigure}[b]{0.49\textwidth}
    \includegraphics[width=\linewidth]{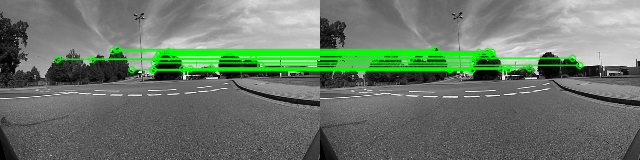}
  \end{subfigure}
    \begin{subfigure}[b]{0.49\textwidth}
    \includegraphics[width=\textwidth]{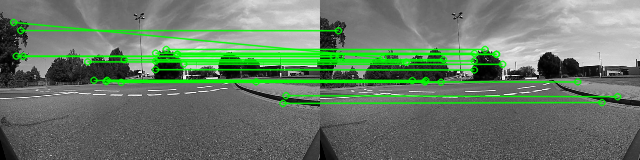}
  \end{subfigure}
  \par\smallskip
    \begin{subfigure}[b]{0.49\textwidth}
    \includegraphics[width=\linewidth]{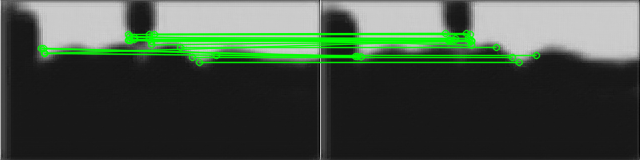}
  \end{subfigure}
    \begin{subfigure}[b]{0.49\textwidth}
    \includegraphics[width=\textwidth]{figs/evaluation/results_orb_pred.png}
  \end{subfigure}
  \caption{Feature matching using ORB (left) and SIFT (right) features. The matched images and common representations are taken at time $t$ and $t+1$, respectively. The left common representation is generated from a grayscale input image, the right one from a projected point cloud depth image.}
  \label{fig:results_matching}
\end{figure}

To demonstrate the usability of the CR, two standard feature matching methods are applied. 
The results are averaged over 100 images from the test data set. 
Figure~\ref{fig:results_matching} shows the result when using ORB and SIFT features for matching. 
The left image pairs are matched using ORB, the right image pairs using SIFT. 
Of each pair, the data is taken of the same scene from different time instances resulting in small variations in the point of view.
The most significant result is the reprojection error given in table~\ref{tab:results_matching}. 
It is significantly higher for the CR in both cases (ORB and SIFT), but still in an acceptable range.
The loss of detail in the CR can be quantified when counting the number of matches that the algorithms find.
Normal images achieve 10 times as many matches as the CR.
On the other hand, the quality of the matches is surprisingly good. 
In fact, the average L2 distance of the matched descriptor vectors is even lower for the CRs, demonstrating the strength of the matches. 

\def\arraystretch{1.5}
\begin{table}[H]
\resizebox{\textwidth}{!}{%
\begin{tabular}{|c|c|c|c|}
\hline
\multicolumn{1}{|c|}{\textbf{Matching inputs}} & \multicolumn{1}{c|}{\textbf{SIFT avg. dist.}} & \multicolumn{1}{c|}{\textbf{SIFT avg. matches}} & \multicolumn{1}{c|}{\textbf{SIFT avg. reprojection error (pixels)}} \\ \hline
image vs. image & 1 & 148 & 0.65 \\ \hline
image CR vs. depth map CR & 0.67 & 13 & 1.46 \\ \hline
\end{tabular}}
\newline
\vspace*{0.2 cm}
\newline
\resizebox{\textwidth}{!}{%
\begin{tabular}{|c|c|c|c|}
\hline
\multicolumn{1}{|c|}{\textbf{Matching inputs}} & \multicolumn{1}{c|}{\textbf{ORB avg. dist.}} & \multicolumn{1}{c|}{\textbf{ORB avg. matches}} & \multicolumn{1}{c|}{\textbf{ORB avg. reprojection error (pixels)}} \\ \hline
image vs. image & 1 & 270 & 1.49 \\ \hline
image CR vs. depth map CR & 0.81 & 23 & 3.68 \\ \hline
\end{tabular}}
\caption{Comparison of image to image matching with common representation matching from RGB and depth image. The distance values of SIFT and ORB feature matches are the L2 norm between matched feature vectors and normalized to 1, to facilitate the interpretation. The number of matches is significantly higher in the image to image matching case, but the resulting reprojection error achieved by the common representation matches is acceptably low.}
\label{tab:results_matching}
\end{table}

\subsection{Architecture 2 - Common Edges}
\label{subsec:results_edges}

\begin{figure}
  \begin{subfigure}{\textwidth}
  \centering
    \includegraphics[width=0.8\textwidth]{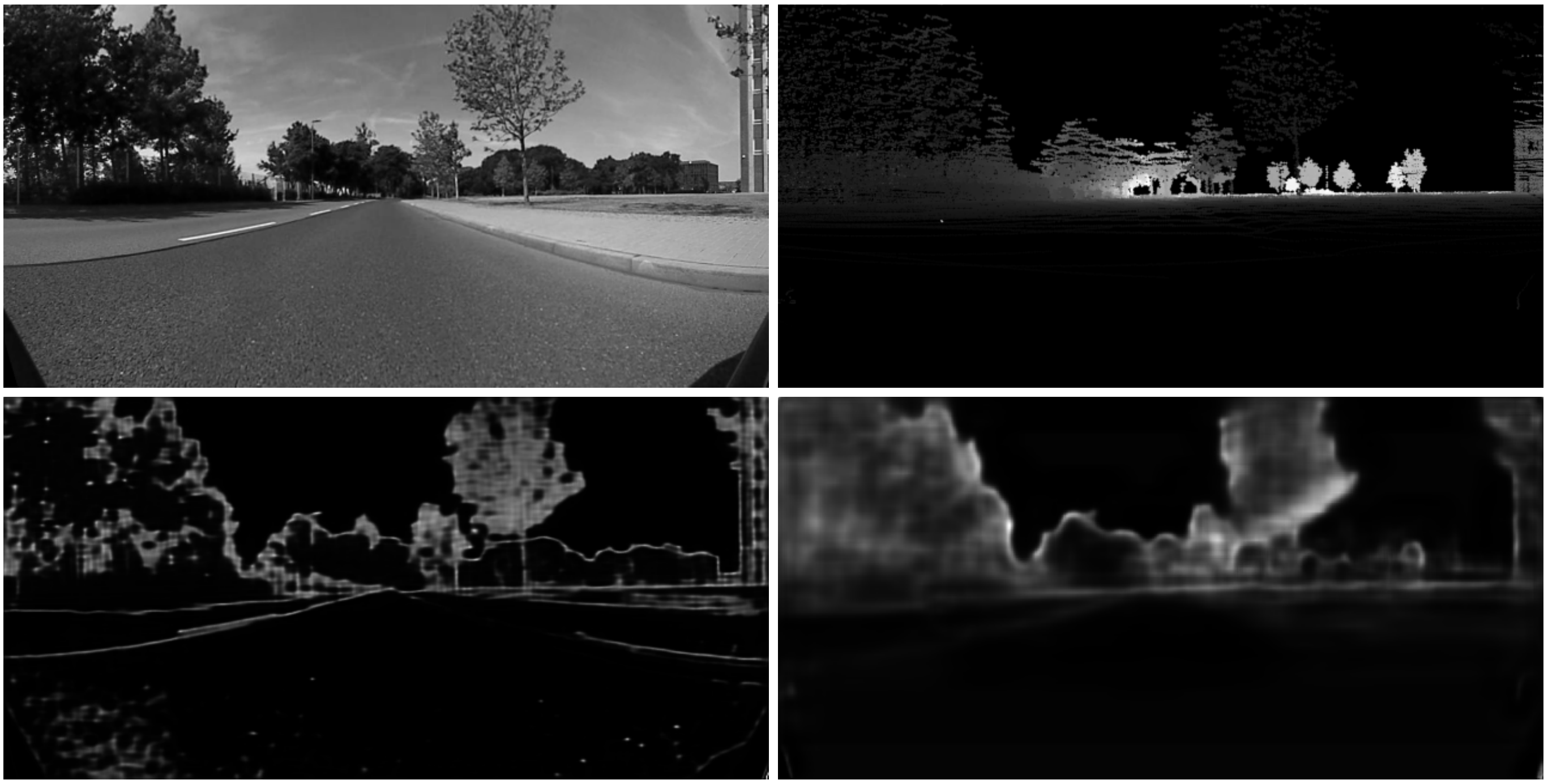}
  \end{subfigure}
  \caption{Top left to bottom right: Input image, input depth image from point cloud, common edges representation generated from input image, common edges representation generated from depth image.}
  \label{fig:results_common_edge}
\end{figure}

The results of the Common Edges approach are depicted in Figure~\ref{fig:results_common_edge}. 
The CR generated from the grayscale image looks similar to an edge image, which agrees with intuition, as the edge image has been generated from the input grayscale image in the first place. 
The CR generated from the depth image is blurry, which does not resemble how an edge image looks like.
More details on this behavior and potential solutions are presented in section~\ref{sec:discussion}.
In general, the results of this architecture are too inconsistent to be used for any matching algorithms. 
A quantitative evaluation is therefore omitted.


\section{Discussion}
\label{sec:discussion}

The results of the Double Siamese Networks show that there is a lack of detail in the common representation, limiting the usability for matching operations. 
Still, the evaluations using ORB and SIFT matches show that the quality of the matches is not too far off the quality of purely image based matches. 

The Common Edges architecture does not perform well at all since the CR generated from the depth image is very blurry.
There are two potential causes for this issue.
Firstly, when utilizing encoder-decoder architectures, it is a common problem that the results look blurry. 
This can potentially be resolved under utilization of GANs, as proposed in~\citep{Sainburg2018GenerativeAI}.
Secondly, there is a conceptual flaw in this approach. 
Namely, many objects have edges in the sense of a Canny edge detector, e.g., red brick buildings, which a depth sensor is not capable to capture. 

The reduction of information when generating a common representation seems to be unavoidable when utilizing an architecture as proposed in this work. 
In order to find a representation that retains more information, it is necessary to incorporate more information in either of the pipelines. 
For instance, one could perform monocular depth estimation from images to reduce the domain gap between camera and LiDAR data. 
A novel approach for this is presented in~\citep{Gaidon18}.

Another approach could include semantic information.
On those lines it could be useful to apply object instance detection on both modalities to generate a higher-level representation of the scene. 
A common method for this is proposed in~\citep{He2017MaskR}, called Mask R-CNN.

The issues with approaches like this are that the generalizability of the whole system depends on the generalization capabilities of the individual parts. 
The training time and the data set size, which is necessary to still be able to generalize well, might not be feasible anymore.


\section{Conclusion}
\label{sec:conclusion}

In this work we have proposed two architectures to learn common representations of LiDAR and camera data, in the form of a 2D image. 
We find that for one of our approaches the results are not as detailed as initially hoped, but that they are still sufficient for basic feature matching algorithms.
The Double Siamese Network architecture showed to be very effective in ensuring that the consistency and distinctiveness of the common representation is given.
We have further shown the limitations of using just geometric data, and propose the incorporation of additional semantic information for future work.



\clearpage


\bibliography{references}  

\begin{thebibliography}{15}
\providecommand{\natexlab}[1]{#1}
\providecommand{\url}[1]{\texttt{#1}}
\expandafter\ifx\csname urlstyle\endcsname\relax
  \providecommand{\doi}[1]{doi: #1}\else
  \providecommand{\doi}{doi: \begingroup \urlstyle{rm}\Url}\fi

\bibitem[{Badrinarayanan} et~al.(2017){Badrinarayanan}, {Kendall}, and
  {Cipolla}]{Badrinarayanan2017}
V.~{Badrinarayanan}, A.~{Kendall}, and R.~{Cipolla}.
\newblock Segnet: A deep convolutional encoder-decoder architecture for image
  segmentation.
\newblock \emph{IEEE Transactions on Pattern Analysis and Machine
  Intelligence}, 39\penalty0 (12):\penalty0 2481--2495, Dec 2017.
\newblock ISSN 0162-8828.
\newblock \doi{10.1109/TPAMI.2016.2644615}.

\bibitem[{Zeng} et~al.(2017){Zeng}, {Yu}, {Wang}, {Li}, and {Tao}]{Zeng2017}
K.~{Zeng}, J.~{Yu}, R.~{Wang}, C.~{Li}, and D.~{Tao}.
\newblock Coupled deep autoencoder for single image super-resolution.
\newblock \emph{IEEE Transactions on Cybernetics}, 47\penalty0 (1):\penalty0
  27--37, Jan 2017.
\newblock ISSN 2168-2267.
\newblock \doi{10.1109/TCYB.2015.2501373}.

\bibitem[Kulkarni et~al.(2015)Kulkarni, Whitney, Kohli, and
  Tenenbaum]{Kulkarni2015}
T.~D. Kulkarni, W.~F. Whitney, P.~Kohli, and J.~Tenenbaum.
\newblock Deep convolutional inverse graphics network.
\newblock In C.~Cortes, N.~D. Lawrence, D.~D. Lee, M.~Sugiyama, and R.~Garnett,
  editors, \emph{Advances in Neural Information Processing Systems 28}, pages
  2539--2547. Curran Associates, Inc., 2015.

\bibitem[Zhu et~al.(2017)Zhu, Park, Isola, and Efros]{Zhu2017}
J.-Y. Zhu, T.~Park, P.~Isola, and A.~A. Efros.
\newblock Unpaired image-to-image translation using cycle-consistent
  adversarial networks.
\newblock In \emph{The IEEE International Conference on Computer Vision
  (ICCV)}, Oct 2017.

\bibitem[Choi et~al.(2017)Choi, Choi, Kim, Ha, Kim, and Choo]{Choi2017}
Y.~Choi, M.~Choi, M.~Kim, J.~Ha, S.~Kim, and J.~Choo.
\newblock Stargan: Unified generative adversarial networks for multi-domain
  image-to-image translation.
\newblock \emph{CoRR}, abs/1711.09020, 2017.
\newblock URL \url{http://arxiv.org/abs/1711.09020}.

\bibitem[Kim et~al.(2017)Kim, Cha, Kim, Lee, and Kim]{Kim2017}
T.~Kim, M.~Cha, H.~Kim, J.~K. Lee, and J.~Kim.
\newblock Learning to discover cross-domain relations with generative
  adversarial networks.
\newblock \emph{CoRR}, abs/1703.05192, 2017.
\newblock URL \url{http://arxiv.org/abs/1703.05192}.

\bibitem[Chung et~al.(2017)Chung, Tahboub, and Delp]{Chung2017}
D.~Chung, K.~Tahboub, and E.~J. Delp.
\newblock A two stream siamese convolutional neural network for person
  re-identification.
\newblock In \emph{The IEEE International Conference on Computer Vision
  (ICCV)}, Oct 2017.

\bibitem[{Melekhov} et~al.(2016){Melekhov}, {Kannala}, and
  {Rahtu}]{Melekhov2016}
I.~{Melekhov}, J.~{Kannala}, and E.~{Rahtu}.
\newblock Siamese network features for image matching.
\newblock In \emph{2016 23rd International Conference on Pattern Recognition
  (ICPR)}, pages 378--383, Dec 2016.
\newblock \doi{10.1109/ICPR.2016.7899663}.

\bibitem[Larsen et~al.(2015)Larsen, S{\o}nderby, and Winther]{Larsen2015}
A.~B.~L. Larsen, S.~K. S{\o}nderby, and O.~Winther.
\newblock Autoencoding beyond pixels using a learned similarity metric.
\newblock \emph{CoRR}, abs/1512.09300, 2015.
\newblock URL \url{http://arxiv.org/abs/1512.09300}.

\bibitem[DeTone et~al.(2017)DeTone, Malisiewicz, and Rabinovich]{DeTone2017}
D.~DeTone, T.~Malisiewicz, and A.~Rabinovich.
\newblock Superpoint: Self-supervised interest point detection and description.
\newblock \emph{CoRR}, abs/1712.07629, 2017.
\newblock URL \url{http://arxiv.org/abs/1712.07629}.

\bibitem[Guo et~al.(2017)Guo, Rana, Ciss{\'{e}}, and van~der Maaten]{Guo2017}
C.~Guo, M.~Rana, M.~Ciss{\'{e}}, and L.~van~der Maaten.
\newblock Countering adversarial images using input transformations.
\newblock \emph{CoRR}, abs/1711.00117, 2017.
\newblock URL \url{http://arxiv.org/abs/1711.00117}.

\bibitem[{Varga} et~al.(2017){Varga}, {Costea}, {Florea}, {Giosan}, and
  {Nedevschi}]{Varga2017}
R.~{Varga}, A.~{Costea}, H.~{Florea}, I.~{Giosan}, and S.~{Nedevschi}.
\newblock Super-sensor for 360-degree environment perception: Point cloud
  segmentation using image features.
\newblock In \emph{2017 IEEE 20th International Conference on Intelligent
  Transportation Systems (ITSC)}, pages 1--8, Oct 2017.
\newblock \doi{10.1109/ITSC.2017.8317846}.

\bibitem[Sainburg et~al.(2018)Sainburg, Thielk, Theilman, Migliori, and
  Gentner]{Sainburg2018GenerativeAI}
T.~Sainburg, M.~Thielk, B.~Theilman, B.~Migliori, and T.~Gentner.
\newblock Generative adversarial interpolative autoencoding: adversarial
  training on latent space interpolations encourage convex latent
  distributions.
\newblock \emph{CoRR}, abs/1807.06650, 2018.

\bibitem[Pillai et~al.(2018)Pillai, Ambrus, and Gaidon]{Gaidon18}
S.~Pillai, R.~Ambrus, and A.~Gaidon.
\newblock Superdepth: Self-supervised, super-resolved monocular depth
  estimation.
\newblock \emph{CoRR}, abs/1810.01849, 2018.
\newblock URL \url{http://arxiv.org/abs/1810.01849}.

\bibitem[He et~al.(2017)He, Gkioxari, Doll{\'a}r, and Girshick]{He2017MaskR}
K.~He, G.~Gkioxari, P.~Doll{\'a}r, and R.~B. Girshick.
\newblock Mask r-cnn.
\newblock \emph{2017 IEEE International Conference on Computer Vision (ICCV)},
  pages 2980--2988, 2017.

\end{thebibliography}

\end{document}